\documentclass[conference]{IEEEtran}
\IEEEoverridecommandlockouts

\usepackage[left=0.75in,
            right=0.75in, 
            top=0.75in,
            bottom=0.75in]{geometry}
            
\usepackage{cite}
\usepackage{amsmath,amssymb,amsfonts}
\usepackage{algorithmic}
\usepackage{graphicx}
\usepackage{rotating}
\usepackage{multirow}
\usepackage{sansmath}
\usepackage{textcomp}
\usepackage{xcolor} 
\def\BibTeX{{\rm B\kern-.05em{\sc i\kern-.025em b}\kern-.08em
    T\kern-.1667em\lower.7ex\hbox{E}\kern-.125emX}}
    
\usepackage[unicode=true,bookmarks=true,bookmarksnumbered=true,bookmarksopen=true, breaklinks=true,pdfborder={0 1 1},backref=page,colorlinks=true]{hyperref}
\usepackage{enumitem}

\begin{document}

\title{Accurate Real-time Polyp Detection in Videos from Concatenation of Latent Features Extracted from Consecutive Frames\\
}

\author{
    \IEEEauthorblockN{Hemin Ali Qadir\textsuperscript{1}, Younghak Shin\textsuperscript{2}, Jacob Bergsland\textsuperscript{1}, Ilangko Balasingham\textsuperscript{1,3}} 
    \vspace{0.3cm}
    \IEEEauthorblockA{\textit{\textsuperscript{1}Intervention Centre, Oslo University Hospital, Oslo, Norway}}
    \IEEEauthorblockA{\textit{\textsuperscript{2}Department of Computer Engineering, Mokpo National University, Mokpo, South Korea}}
    \IEEEauthorblockA{\textit{\textsuperscript{3}Department of Electronic Systems, Norwegian University of Science and Technology, Trondheim, Norway}}
}

\maketitle
\thispagestyle{plain}
\pagestyle{plain}

\begin{abstract}
An efficient deep learning model that can be implemented in real-time for polyp detection is crucial to reducing polyp miss-rate during screening procedures. Convolutional neural networks (CNNs) are vulnerable to small changes in the input image. A CNN-based model may miss the same polyp appearing in a series of consecutive frames and produce unsubtle detection output due to changes in camera pose, lighting condition, light reflection, etc. In this study, we attempt to tackle this problem by integrating temporal information among neighboring frames. We propose an efficient feature concatenation method for a CNN-based encoder-decoder model without adding complexity to the model. The proposed method incorporates extracted feature maps of previous frames to detect polyps in the current frame. The experimental results demonstrate that the proposed method of feature concatenation improves the overall performance of automatic polyp detection in videos. The following results are obtained on a public video dataset: sensitivity 90.94\%, precision 90.53\%, and specificity 92.46\%.  
\end{abstract}


\section{Introduction}
Colorectal cancer ranks third in terms of worldwide incidence, but second in terms of mortality for both genders \cite{sung2021global}. Most cases of colorectal cancer originate from abnormal growths of glandular tissue in the inner lining of the colon and rectum. These abnormal tissue growths are known as polyps which are benign in the early stage. Untreated polyps might become malignant and potentially life-threatening cancer \cite{gschwantler2002high}. Colonoscopy is the gold standard method for colon screening and allows the detection and removal of polyps during the procedure. It has been reported that polyp miss rate can be as high as 22\%-28\% depending on the experience of the endoscopists \cite{leufkens2012factors}.

During the last few years, deep learning (DL), a type of machine learning and artificial intelligence (AI), has proven to be the most successful computational method for automatic polyp detection and reduction of polyp miss rate. In particular, convolutional neural networks (CNN), a special form of DL applied for image analysis, have shown outstanding performance in automatic polyp detection and segmentation in colonoscopy images and videos \cite{wang2022afp, yang2022automatic, qadir2021toward, qadir2019improving, angermann2017towards}. Several studies, however, demonstrated that DL-based networks including CNNs are vulnerable to perturbations and noise \cite{moosavi2017universal_attack,narodytska2017simple_attack,nguyen2015deep_attack,papernot2017practical_attck,moosavi2016deepfool,su2017one_pixel}. Jiawei Su et al. \cite{su2017one_pixel} showed that CNNs can be easily fooled by small attacks e.g. by adding relatively small perturbations (one pixel) to the input image. In colonoscopy video analysis, CNNs might be fooled by the specular light reflections and small changes in polyp (other elements) structures appearance. This means that CNNs can easily miss the same polyp presenting in a sequence of consecutive frames and produce unstable detection output contaminated with a high number of false positives and false negatives. 

In this paper, we propose a novel method to address the above-mentioned problem. The proposed algorithm concatenates the features maps of previous frames with the current frame at the bottleneck layer (latent space) of an encoder-decoder based CNN architecture without adding too much complexity. The hypothesis is that neighboring frames are closely related to each other, and thus their extracted CNN features should be closely similar and contain complementary information. We choose to integrate the proposed method into a two-dimensional (2D) CNN-based encoder-decoder network because its elegant architecture facilitates the concatenation of features extracted from a series of consecutive frames by the encoder part in the latent space. Furthermore, most of the 2D CNN-based encoder-decoder networks are designed from fully convolutional neural networks (F-CNN) predicting outputs in a single shot feed-forward manner which makes them eligible for real-time implementation. Like \cite{qadir2021toward}, we enforce the decoder part to predict 2D Gaussian shapes for polyp regions presented in the current input frames from the concatenated features. We demonstrate that the proposed method is efficient to increase polyp detection capability by increasing the number of true positives and reducing the number of false positives.

\begin{figure*}[!t]
    \centering
    \includegraphics[scale=0.37]{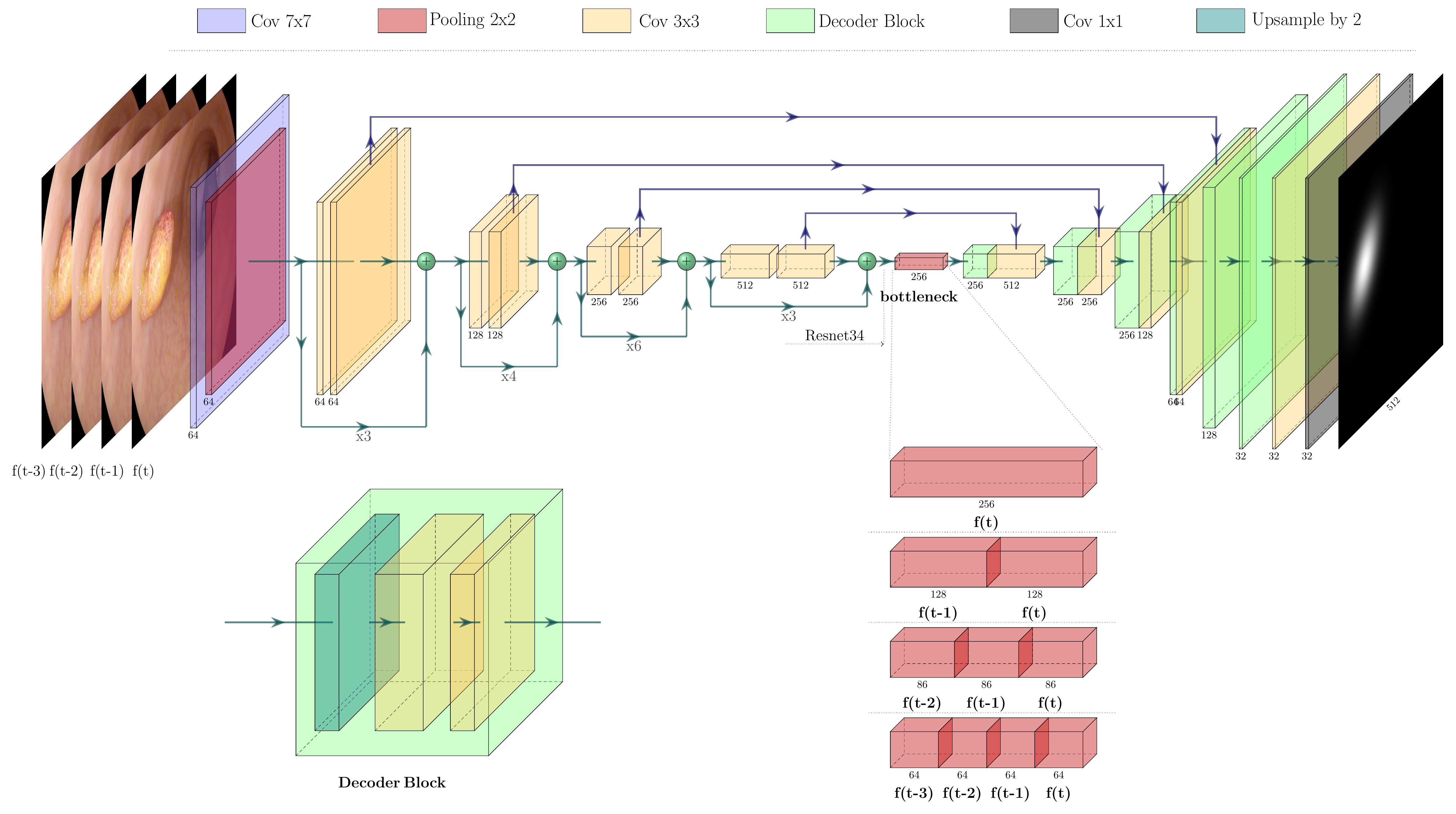}
    \caption{The proposed model which consists of two paths: an encoder extracting features from the input frames, and a decoder interpreting the extracted features to predict the final detection output.}
    \label{fig:model}
\end{figure*}

\section{Materials and Methods}
\subsection{Methodology}
Fig. \ref{fig:model} presents our proposed method to detect polyps in a one-shot manner in videos. The method is developed based on a 2D CNN encoder-decoder network. The 2D encoder-decoder networks are originally developed for single image analysis but do not incorporate temporal information among neighboring frames when they are applied for video analysis. This property together with the CNN vulnerability make the 2D encoder-decoder networks produce unstable output predictions for consecutive frames contaminated with a lot of false detection outputs. In this section, we provide a detailed description of our proposed method to make a CNN-based encoder-decoder network a more stable and reliable polyp detector suitable for video analysis. We incorporate temporal information from previous frames to analyze the current frame. We concatenate extracted features from $N$ previous frames $f_{xy}(t-n), n \in [1,2, ..N]$ with the extracted features of the current frame $f_{xy}(t)$ in the bottleneck layer (latent space). 

\vspace{1em}
\subsubsection{Network architecture}
In this work, we adapt AlbuNet34 proposed by Shvets et al. \cite{shvets2018angiodysplasia} for our polyp detection model as shown in Fig. \ref{fig:model}. AlbuNet34 is a UNet-like architecture \cite{ronneberger2015u} consisting of two paths: the contracting path and expanding path. The contracting path (encoder part) takes in an input image frame $f_{xy}(t)$ and progressively extracts abstract features. The expansive path (decoder part) interprets the extracted features and enables precise localization. AlbuNet34 uses ResNet34 \cite{he2016deep} pre-trained on Imagenet dataset \cite{deng2009imagenet} for the encoder part. We choose AlbuNet34 because it combines the advantages of both UNet-like architectures \cite{ronneberger2015u} and residual learning \cite{he2016deep}.  In addition, AlbuNet34 is designed from fully convolutional neural networks (F-CNN) predicting outputs in a single shot feed-forward manner which makes them eligible for real-time implementation.

The first block of the encoder is a kernel of size $7 \times 7$ with stride 2 followed by a max-pooling layer with stride 2. The rest blocks of the encoder consist of repetitive residual blocks. In every residual block, the first convolution operation is applied with stride 2 to provide downsampling, while the rest convolution operations are applied with stride 1. We apply a $2 \times 2$ max-pooling operation on the final output feature maps of the final residual block of ResNet34. The result of this max-pooling operation is stored in the bottleneck layer (the latent space). We add this bottleneck layer to facilitate the incorporation of temporal information from consecutive frames, which will be discussed in Section \ref{sec:concat}. 

The decoder part consists of several decoder blocks, each block is concatenated with the corresponding encoder block. In every decoder block, an upsampling operation is applied to upsample the feature maps by 2, followed by two padded convolution operations of a kernel of size $3 \times 3$ with stride 1 followed by a rectified linear unit (ReLU). The first block of the decoder starts with interpreting the abstract features stored in the bottleneck layer. To generate the final output, we apply a $1 \times 1$ convolution operation followed by the $tanh$ activation function. Our detection model generates an output image which has the same resolution as the input image frame $f_{xy}(t)$, with the predicted 2D Gaussian shapes $\hat{Y}_{xy}(t)$ discussed in Section \ref{sec:2d_gaussian}.

\vspace{1em}
\subsubsection{Concatenation of consecutive features}
\label{sec:concat}
Qadir et al. \cite{qadir2019improving} showed that the same CNN-based detector can miss the same polyp appearing in the neighboring frames due to changes in the light conditions, appearances, inherent noises, blurriness, etc. This is due to the vulnerability of CNN to small perturbations. We solve this problem by concatenating the feature maps of a series of consecutive frames extracted by the encoder part. Neighboring frames are closely related to each other and thus their extracted feature maps should be closely similar and contain complementary information. The elegant structure of UNet-like architectures facilitates this feature concatenation in the bottleneck layer. This way we can incorporate temporal information among neighboring frames into the detection model.  

We store the feature maps from $N$ previous frames $f_{xy}(t-n), n \in [1,2, ..N]$ and concatenate them with the feature maps extracted from the current frame $f_{xy}(t)$ in the bottleneck layer. We set the number of activation maps in the bottleneck layer to be 256 maps. We equally divide the bottleneck layer into $N$ slots of $256/N$ activation maps based on the number of previous frames involved. For instance, when only one previous frame $f_{xy}(t-1)$ is incorporated, 128 feature maps are extracted from the current frame $f_{xy}(t)$ and 128 feature maps from $f_{xy}(t-1)$. However, when the result of this division is a floating number we round it up to an approximate number. For instance, when we consider two previous frames $f_{xy}(t-n), n \in [1,2]$, the result of 256/3 is 85.33, thus we use 86 maps for each frame, resulting in 258 feature maps in the bottleneck layer.  

This concatenation of feature maps helps combine complementary information from a series of previous frames  $f_{xy}(t-n), n \in [1,2, ..N]$ with the current frame $f_{xy}(t)$, reducing the effect of a small perturbation and/or change that may pop up in the current frame $f_{xy}(t)$ and fool the CNN-based detector. Therefore, this concatenation strategy helps the CNN-based encoder-decoder network improve its accuracy and produce more stable detection outputs for a series of neighboring frames.

\vspace{1em}

\subsubsection{2D Gaussian shapes for polyp regions} 
\label{sec:2d_gaussian}
Qadir et al. \cite{qadir2021toward} demonstrated that a CNN-based encoder-decoder network is more efficient in detecting polyps when it is trained on 2D Gaussian shapes as the ground-truth masks instated of using binary masks. We follow the same procedure proposed in \cite{qadir2021toward} to train our detection model to predict a 2D Gaussian shape, $\hat{Y}_{xy}(t) \in [0,1]^{W \times H \times 1}$, for a polyp region in an input RGB image frame at time $t$, $f_{xy}(t) \in [R]^{W \times H \times 3}$, where $W$ is the width and $H$ is the height of both $f_{xy}(t)$ and $\hat{Y}_{xy}(t)$. 

We transform the provided binary ground-truth masks, $X_{xy}(t) \in\{0,1\}^{W \times H \times 1}$, to 2D Gaussian ground-truth masks, $Y_{xy}(t) \in [0,1]^{W \times H \times 1}$, as described in \cite{qadir2021toward}. The 2D Gaussian ground-truth masks are meant to reduce the impact of the outer edges during training and force the network to learn the surface patterns of different polyps more efficiently.  

Using 2D Gaussian shapes can also help to realize polyp detection in real-time. We can generate all detected bounding boxes directly from the predicted 2D Gaussian shapes without the need for computing non-maximum suppression (NMS) to eliminate overlapping bounding boxes \cite{zhou2019objects}. At the inference time, we use the strength of the predicated 2D Gaussian shapes as the confidence values of the detected bounding boxes and calculate the two size-adaptive standard deviations ($\sigma_x$ and $\sigma_y$) for the size of the detected bounding boxes.

\subsection{Datasets}
\label{sec:label}
In this study, we used four publicly available datasets of still images and videos: 
\begin{enumerate}[label=\alph*)]
    \item CVC-ColonDB \cite{bernal2012towards}: This is a dataset of 300 still images extracted from 15 colonoscopy videos, each with a unique polyp (15 unique polyps in total). The images have a resolution of 574x500 pixels. 
    \item CVC-ClinicDB \cite{bernal2015wm}: This is a dataset of 612 still images extracted from 29 colonoscopy videos, each with at least a polyp. There exists 31 unique polyps presented 646 times in the 612 images with a pixel resolution of 384x288.  
    \item ASU-Mayo Clinic \cite{tajbakhsh2015automated}: This is a dataset of 38 colonoscopy videos. 20 videos are assigned as a training set while the other 18 videos are assigned as a testing set. Because of the copyright license, we could not get access to the 18 testing videos. The 20 training videos consist of 10 positive videos with a total of 5402 frames (3846 polyp frames) and 10 negative videos with a total of 13500 frames.
    \item CVC-ClinicVideoDB \cite{angermann2017towards}: This is a dataset of 18 colonoscopy videos, each with a different polyp. This dataset comprises 11954 frames (10025 polyp frames). The resolution of the frames is 268x576 pixels.
\end{enumerate}

\subsection{Training Details}
During training, a video dataset is needed in order to capture the temporal patterns among neighboring frames. In a video, the neighboring frames look closely similar. If there are not enough diverse frames in the training videos, our detection model may easily get overfitted on the training frames. To train our detection model and avoid this phenomenon, we use the two datasets of still images namely CVC-ColonDB and CVC-ClinicDB alongside the dataset of videos namely CVC-ClinicVideoDB. We build our final training dataset by mixing the frames of the videos and the still images. Whenever a still image is encountered during training, we count it as its previous frames as well.

\begin{figure}[!ht]
    \centering
    \begin{tabular}{cc}
         \includegraphics[height=3.5cm, width=4.5cm]{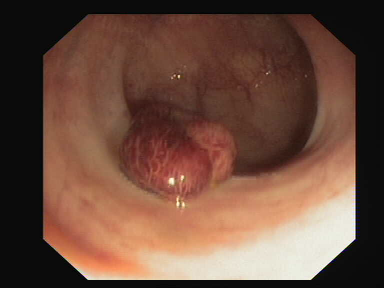} & \includegraphics[height=3.5cm, width=3.5cm]{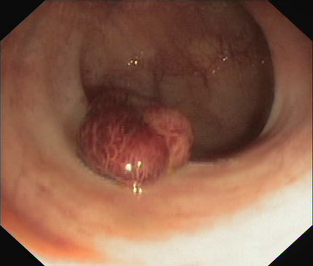} \\
         \footnotesize (a) The original image & \footnotesize (b) Cropped and resized image 
    \end{tabular}
    \caption{An example showing image cropping and resizing to $512 \times 512$. a) }
    \label{fig:cropping}
\end{figure}

As mentioned before, the encoder part uses ResNet34 initialized with ImageNet pre-trained weights. In contrast, we randomly initialize the network parameters of the decoder part. To clean the final training dataset, we apply several simple pre-processing methods to the input images (see Fig. \ref{fig:cropping}): 
\begin{enumerate}
    \item Image cropping: to remove the canvas around the informative part of the images. 
    \item Image resizing: by changing the image resolution to $512 \times 512$ because the pre-trained ResNet34 accepts this image resolution. 
    \item Image normalization: by converting the pixel values from [0, 255] to [0, 1], subtracting them from the mean, and dividing them by standard deviation both pre-calculated from the ImageNet dataset. 
\end{enumerate}

To add father image-level diversity through depth and scale, we apply several image augmentation methods on the fly e.g. rotation, vertical, horizontal flips, random zoom-in (up to 25\%), and zoom-out (up to 50\%), and color augmentations in HSV space. To keep the balance between large and small polyps and avoid biasing, we apply less zoom-in compared to zoom-out because the training dataset contains more large polyps than small ones. 

We randomly split the training dataset into training (85\%) and validation (15\%) subsets. We use Adam optimizer with a batch size of 10 and a learning rate of $1 \times 10^{-4}$ to train the model for 20 epochs. Following the recommendations given in \cite{shvets2018angiodysplasia}, we change the learning rate to $1 \times 10^{-5}$ to train the model up to 60 epochs. We use the validation subset to choose the learning rate decay strategy and the number of epochs.  

Finally, we use the mean squared error (squared L2 norm) between each element in the input ground-truth image frame at time $t$, $Y_{xy}(t)$, and the output image frame at time $t$, $\hat{Y}_{xy}(t)$ with the predicted 2D Gaussian shapes.

\begin{equation}
    L2\: loss = \frac{1}{M} \sum^M_i [Y_{xy}(t)_i - \hat{Y}_{xy}(t)_i]^2,
\end{equation} 
where M is the batch size. We choose L2 norm loss function because it can significantly (quadratically) penalize large errors. This property of L2 norm makes it favorable especially for the prediction of 2D Gaussian shapes which are normally distributed around a mean value.

\subsection{Evaluation Metrics}
The output of the proposed method is a set of bounding boxes around the suspected regions in the input frame. The detected bounding boxes are either true or false alarms. To quantitatively evaluate the performance of the proposed method, we calculate sensitivity (recall) and precision using well-known medical parameters:

\begin{enumerate}[label=\alph*)]
\item true positive (TP): a true detected bounding box around a positive region in the input frame,
\item false positive (FP): a false detected bounding box around a negative region,
\item true negative (TN): a true detection output for a negative frame in which no bounding box is detected. 
\item false negative (FN): false detection output where a polyp is missed in a positive frame. 
\end{enumerate}
Sensitivity measures the ratio of TPs to the total number of polyps in the test set,
\begin{equation}
    Sensitivity = TP/(TP+FN) \times 100,
\end{equation}
while precision measures the ratio of TPs to the total number of detected bounding boxes including FPs,
\begin{equation}
    Precision = TP/(TP+FP) \times 100,
\end{equation}
specificity measures the ratio of actual negative frames that are correctly classified,
\begin{equation}
    Specificity = TN/(TN+FP) \times 100.
\end{equation}

\section{Results and Discussion}
To quantitatively evaluate our proposed method, we used the ASU-Mayo clinic dataset, more specifically the 20 videos that were originally assigned for training purposes by the authors. The 10 positive videos were used to compute the performance of the proposed method in terms of sensitivity and precision. In contrast, the 10 negative videos were used for the evaluation of specificity. We present our results in curves to facilitate the visualization of the performance evaluation when information from previous frames is incorporated with the current frame.

\subsection{Results on positive videos}
\begin{figure}[htbp]
\centerline{\includegraphics[scale=0.53]{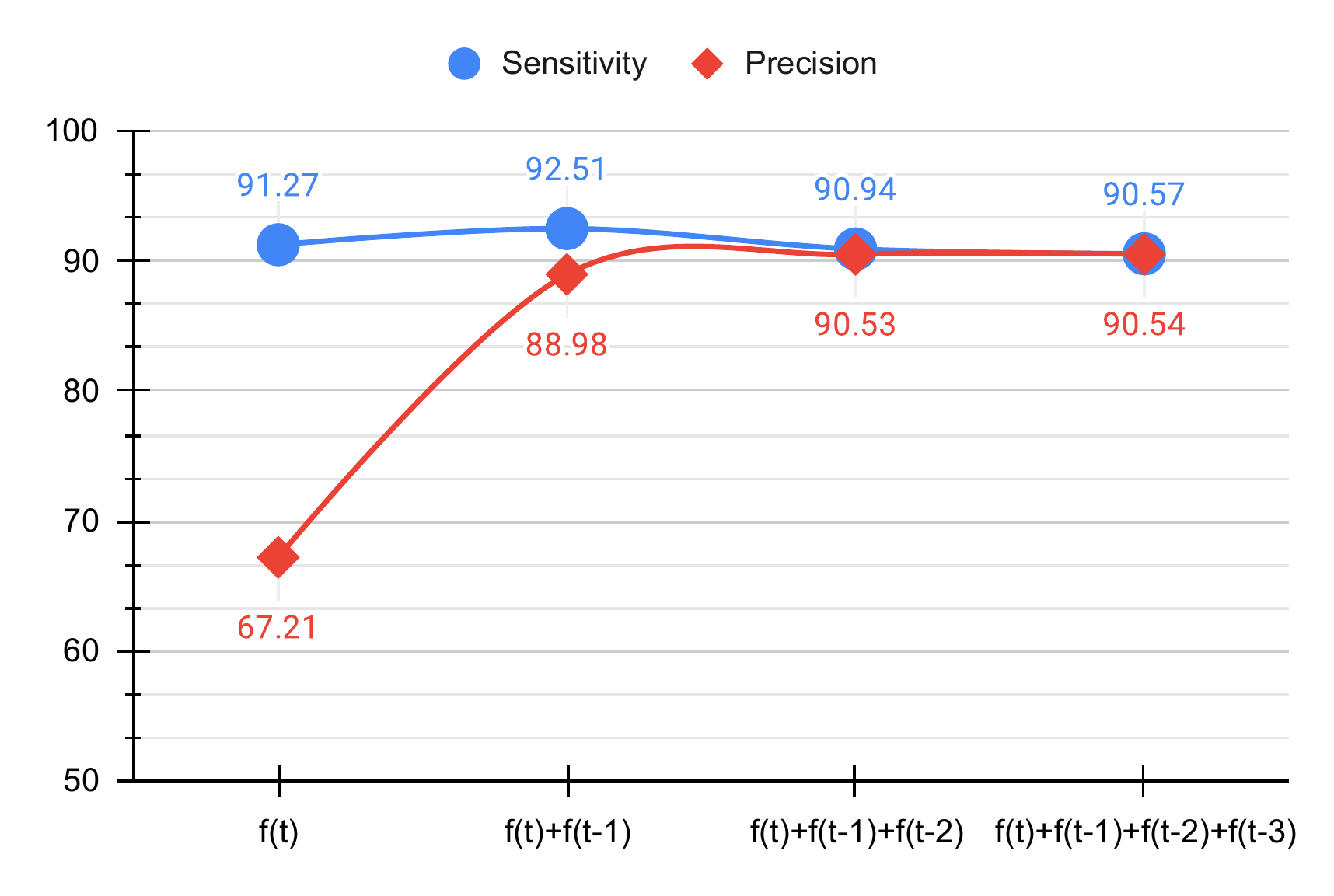}}
\caption{Numeral results showing the performance improvement of the proposed method in terms of sensitivity and precision}
\label{fig:SenPre}
\end{figure}

Fig. \ref{fig:SenPre} shows sensitivity and precision measurement of the proposed method for all four scenarios. When the current frame is examined alone, AlbuNet34 can provide high sensitivity (91.27\%) but struggles to offer the same level of performance for precision (67.21\%) due to the generation of a substantial number of FPs.

\begin{figure*}[!t]
\begin{centering}
{\footnotesize{}%
\begin{tabular}{l}

\includegraphics[width=0.8in,height=0.7in]{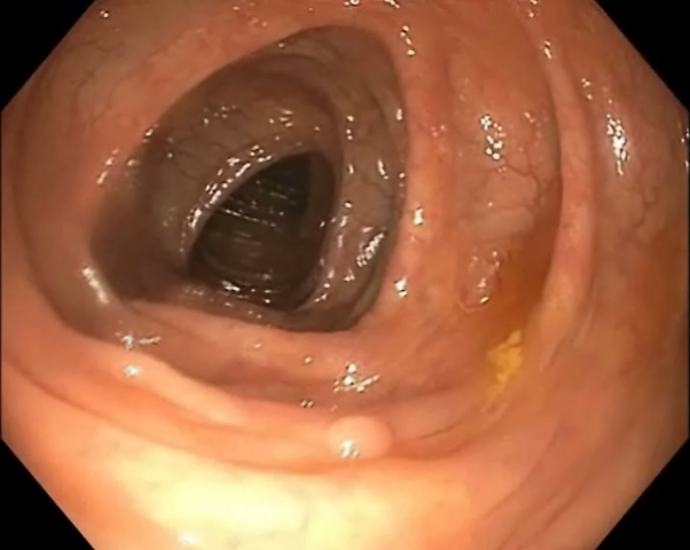}
\includegraphics[width=0.8in,height=0.7in]{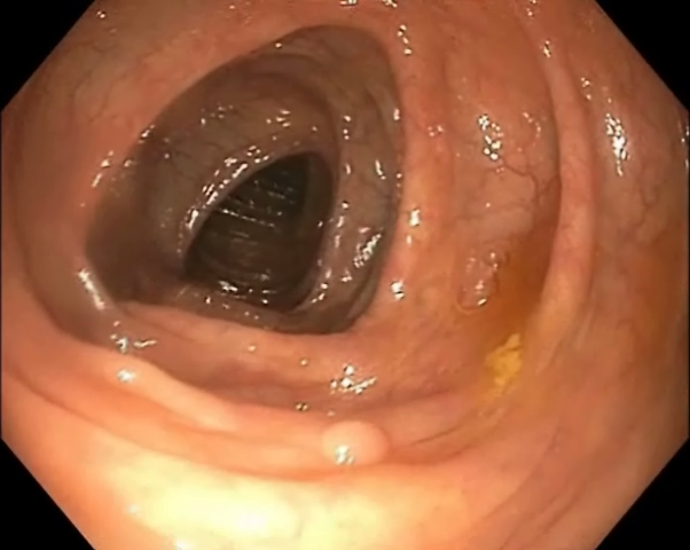}
\includegraphics[width=0.8in,height=0.7in]{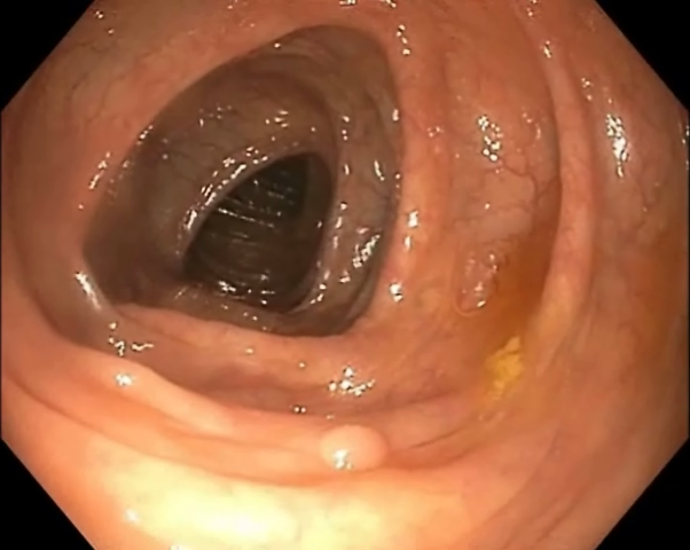}
\includegraphics[width=0.8in,height=0.7in]{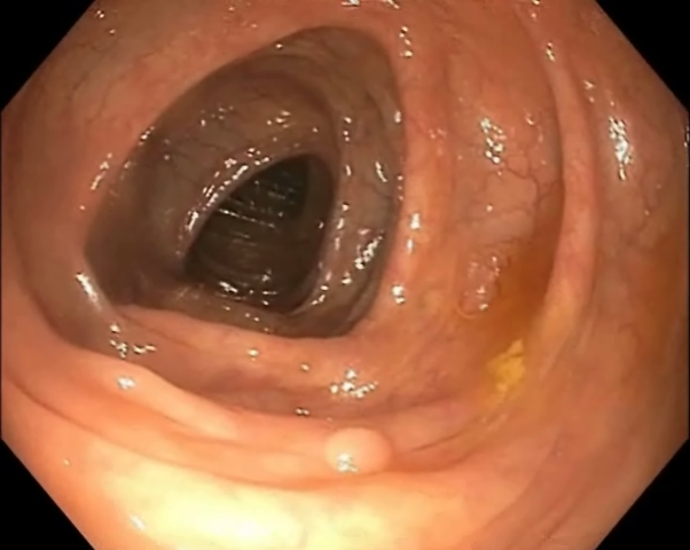}
\includegraphics[width=0.8in,height=0.7in]{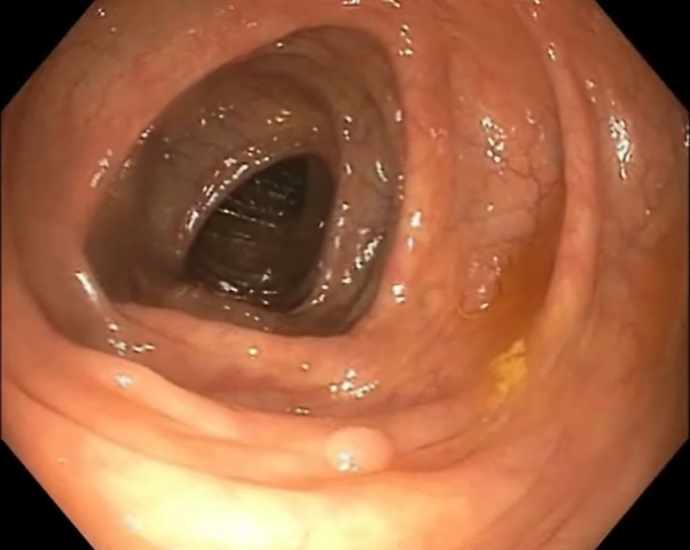}
\includegraphics[width=0.8in,height=0.7in]{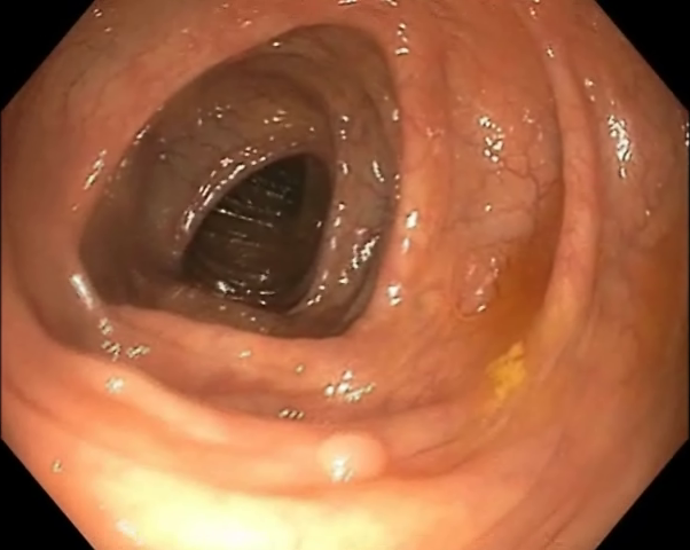}
\includegraphics[width=0.8in,height=0.7in]{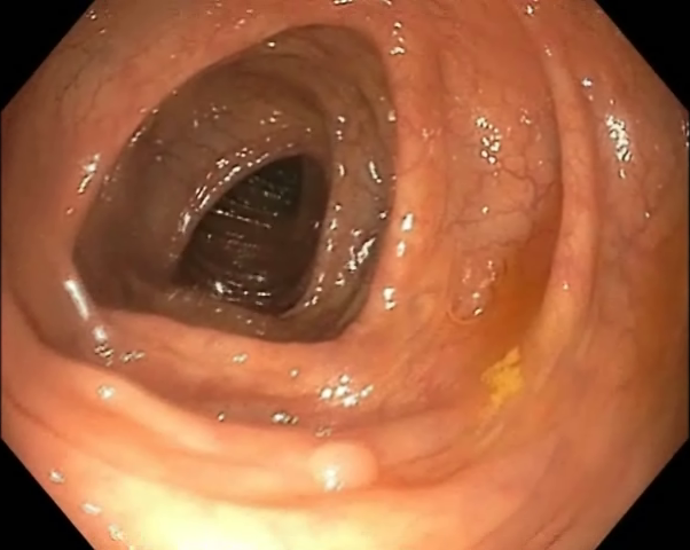}
\includegraphics[width=0.8in,height=0.7in]{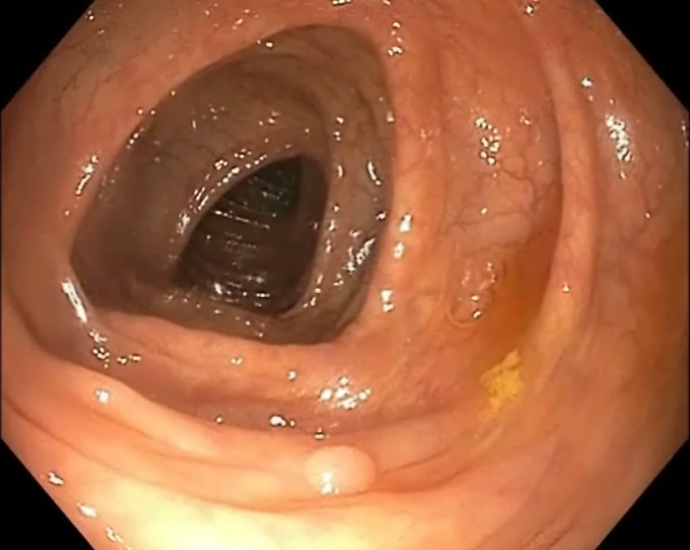}\tabularnewline

\includegraphics[width=0.8in,height=0.7in]{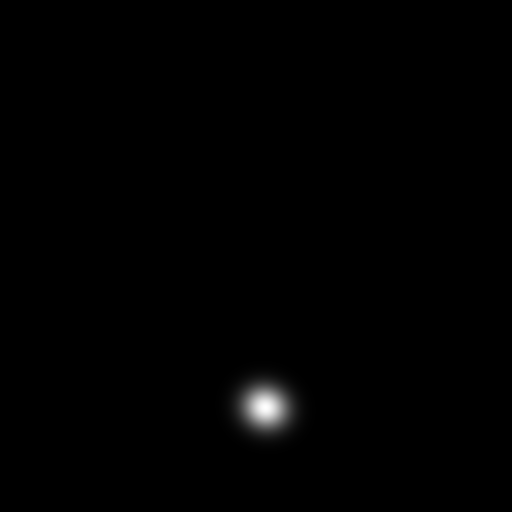}
\includegraphics[width=0.8in,height=0.7in]{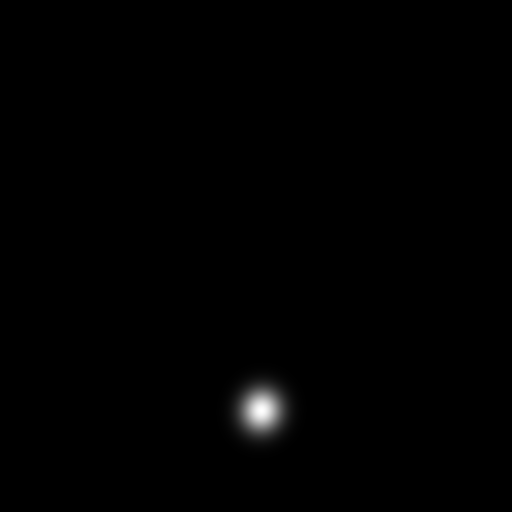}
\includegraphics[width=0.8in,height=0.7in]{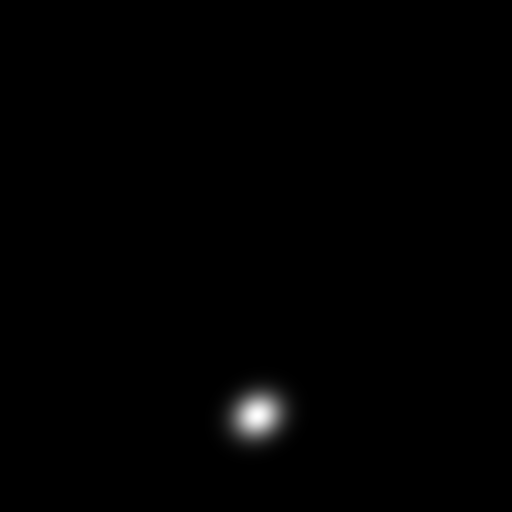}
\includegraphics[width=0.8in,height=0.7in]{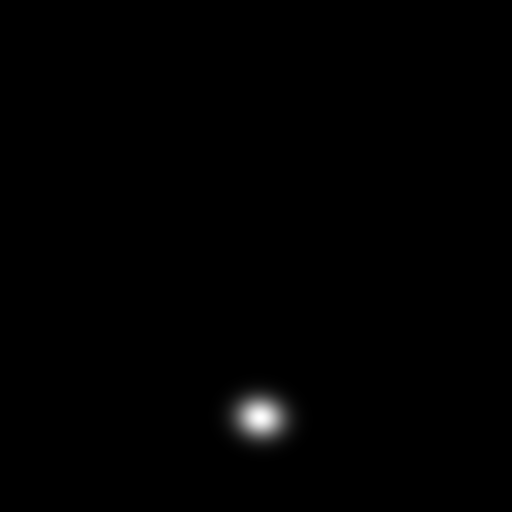}
\includegraphics[width=0.8in,height=0.7in]{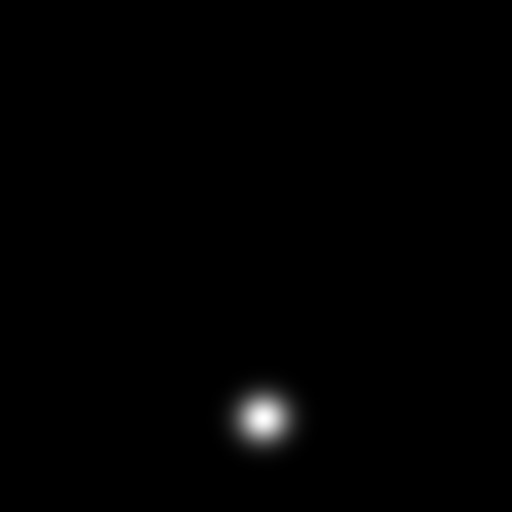}
\includegraphics[width=0.8in,height=0.7in]{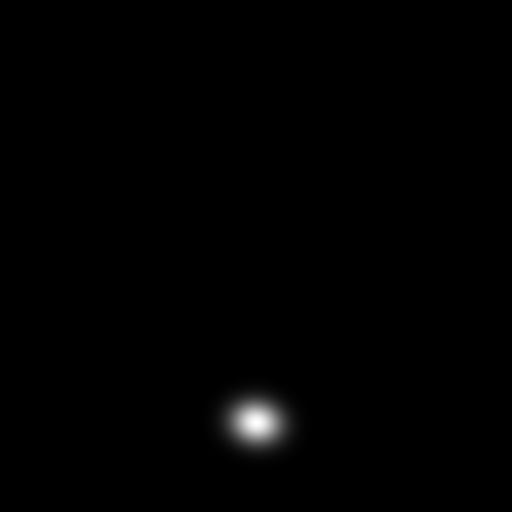}
\includegraphics[width=0.8in,height=0.7in]{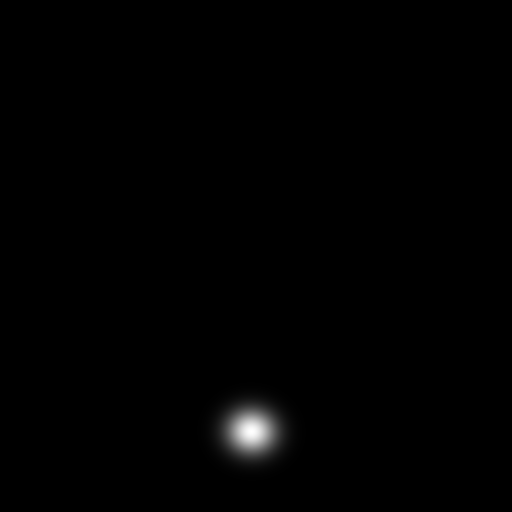}
\includegraphics[width=0.8in,height=0.7in]{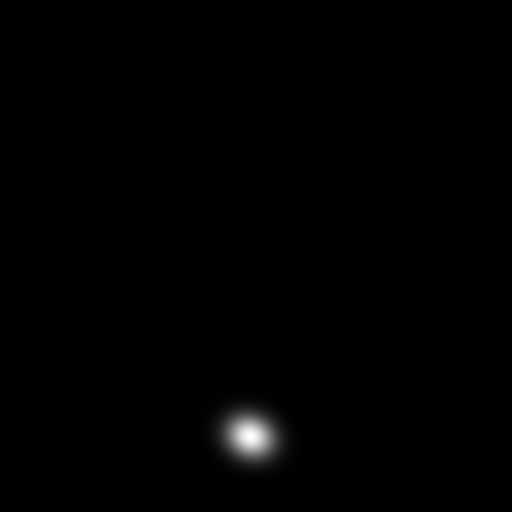}\tabularnewline

\includegraphics[width=0.8in,height=0.7in]{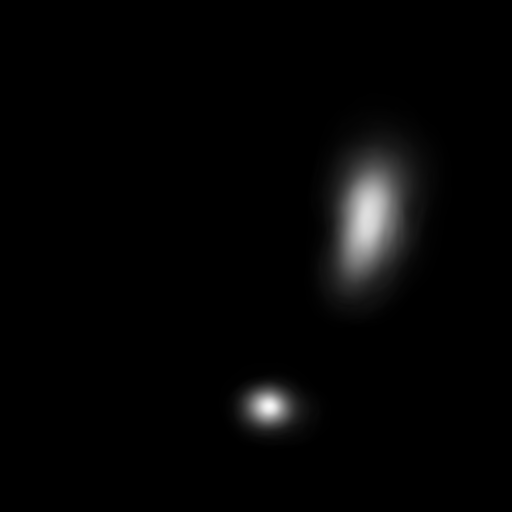}
\includegraphics[width=0.8in,height=0.7in]{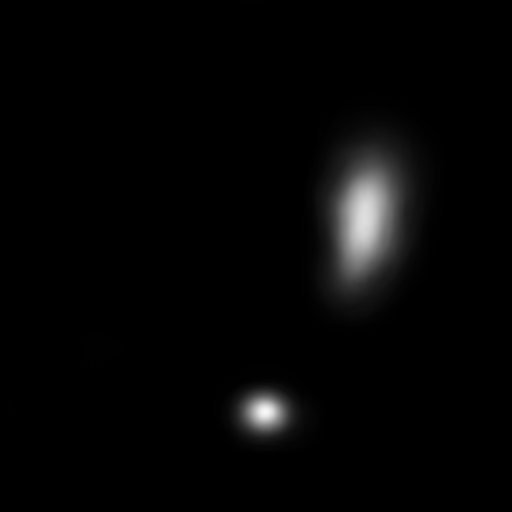}
\includegraphics[width=0.8in,height=0.7in]{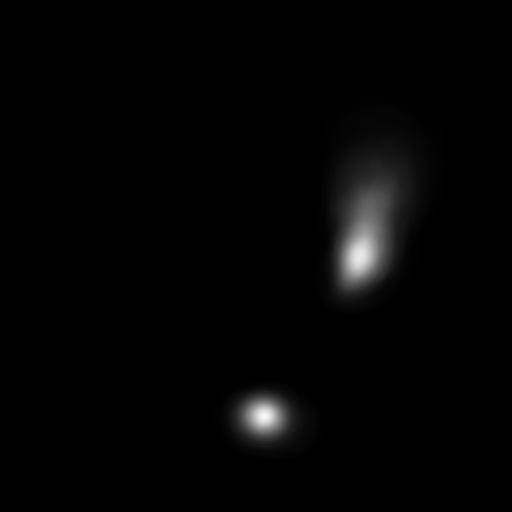}
\includegraphics[width=0.8in,height=0.7in]{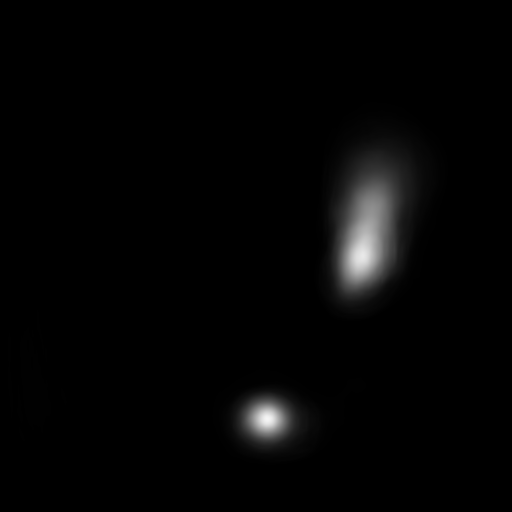}
\includegraphics[width=0.8in,height=0.7in]{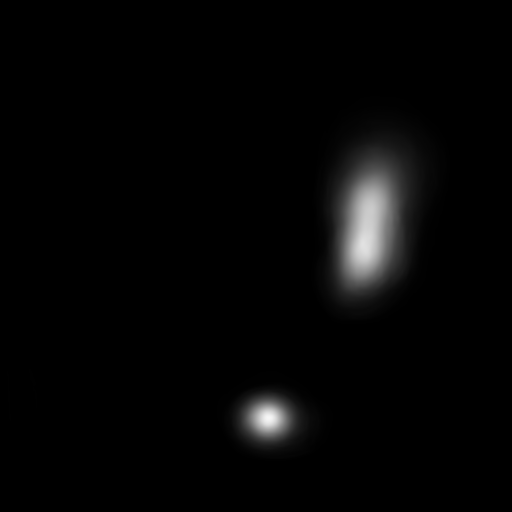}
\includegraphics[width=0.8in,height=0.7in]{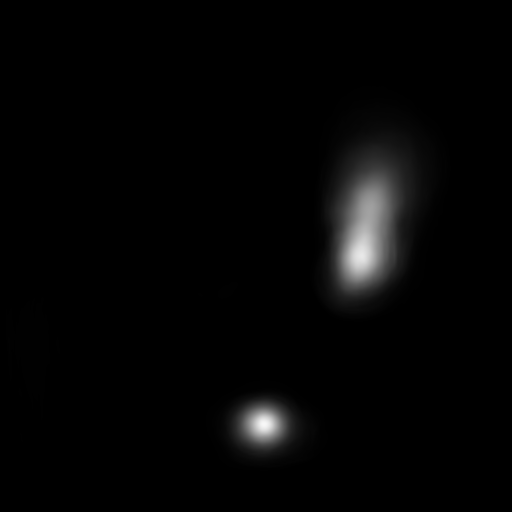}
\includegraphics[width=0.8in,height=0.7in]{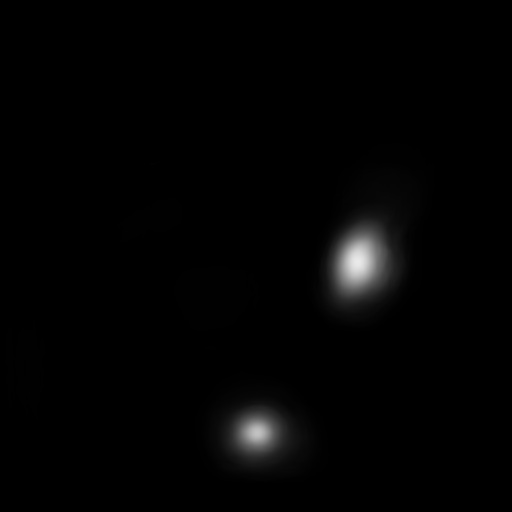}
\includegraphics[width=0.8in,height=0.7in]{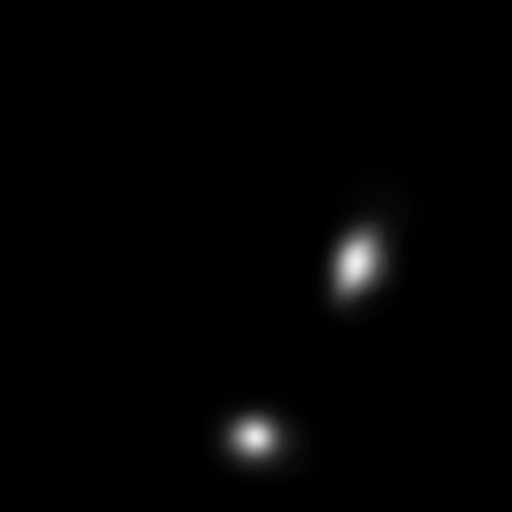}\tabularnewline

\includegraphics[width=0.8in,height=0.7in]{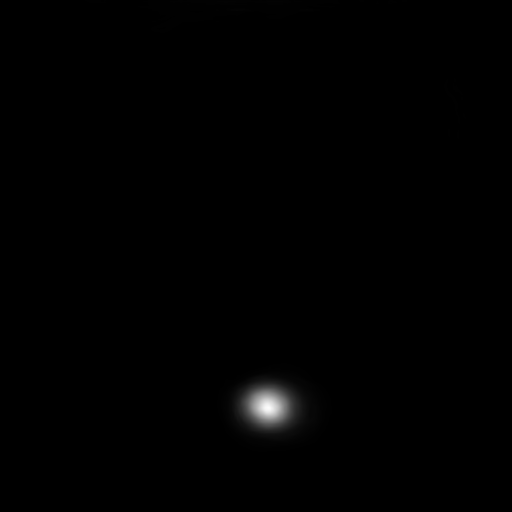}
\includegraphics[width=0.8in,height=0.7in]{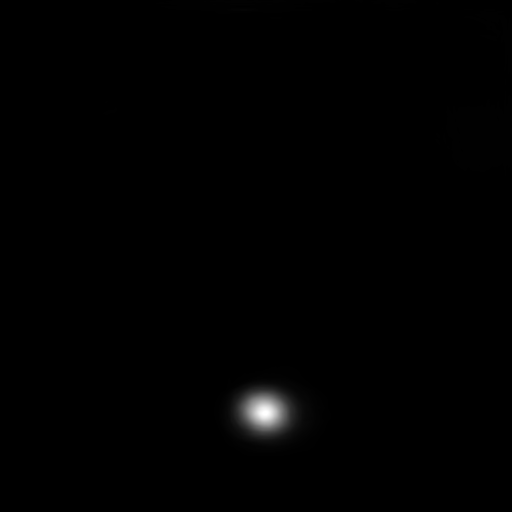}
\includegraphics[width=0.8in,height=0.7in]{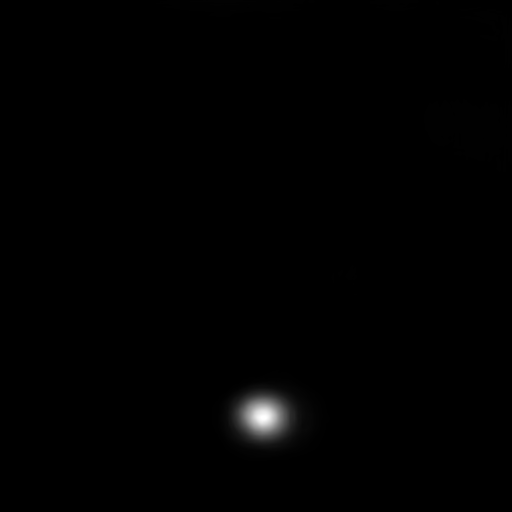}
\includegraphics[width=0.8in,height=0.7in]{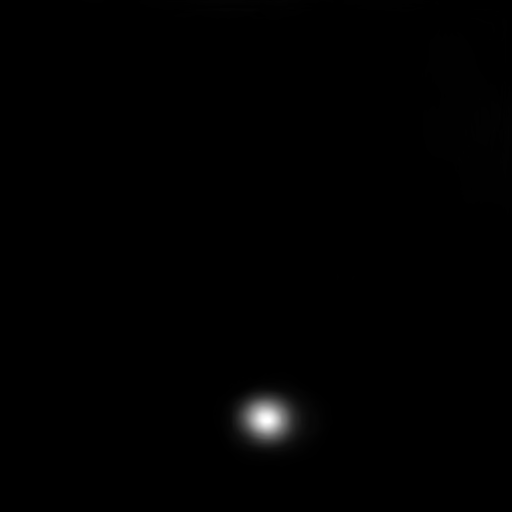}
\includegraphics[width=0.8in,height=0.7in]{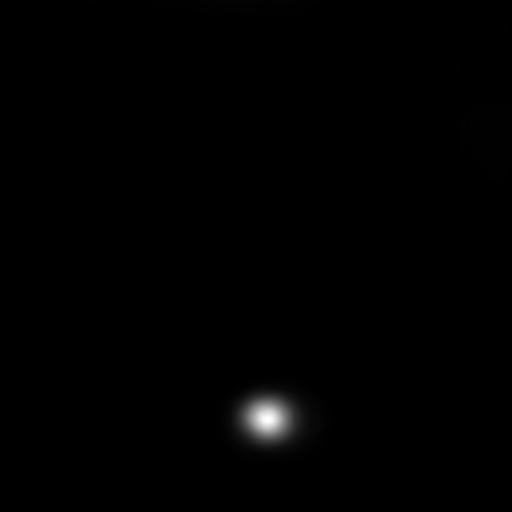}
\includegraphics[width=0.8in,height=0.7in]{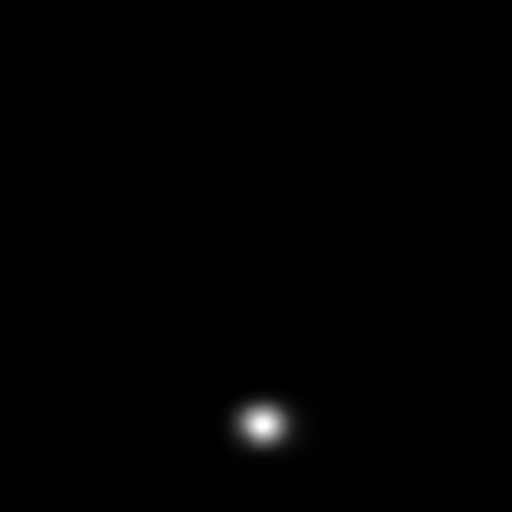}
\includegraphics[width=0.8in,height=0.7in]{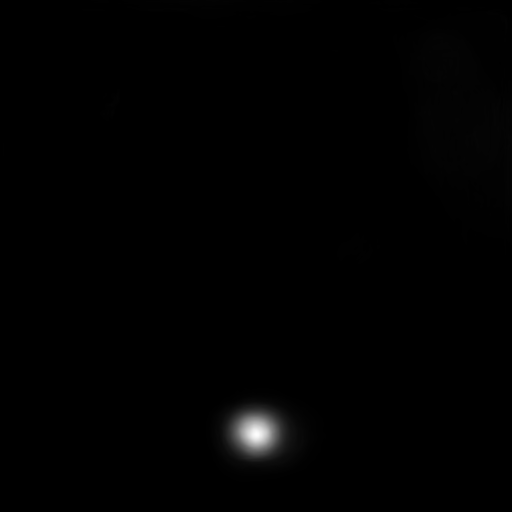}
\includegraphics[width=0.8in,height=0.7in]{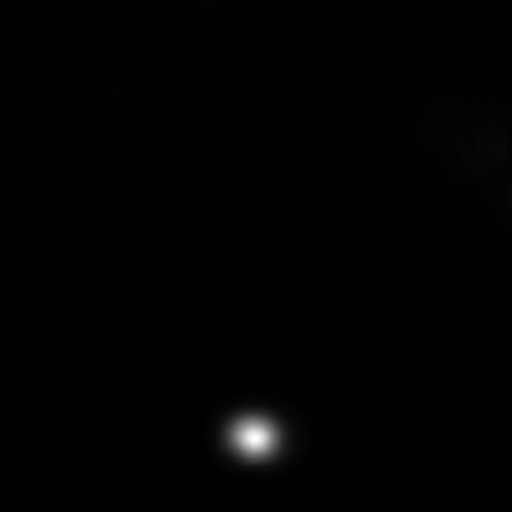}\tabularnewline

\includegraphics[width=0.8in,height=0.7in]{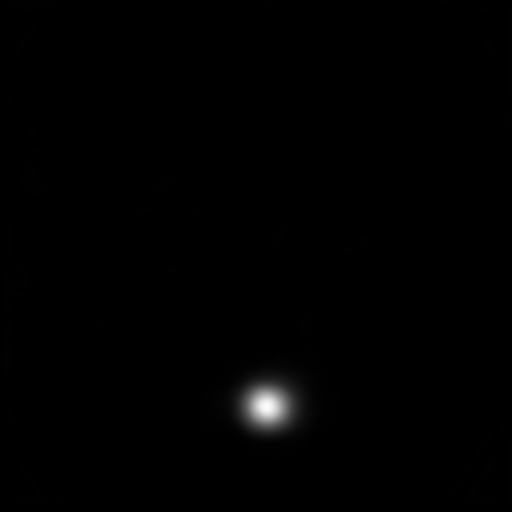}
\includegraphics[width=0.8in,height=0.7in]{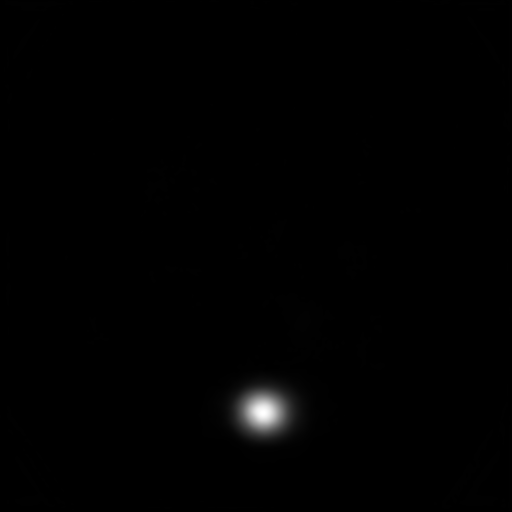}
\includegraphics[width=0.8in,height=0.7in]{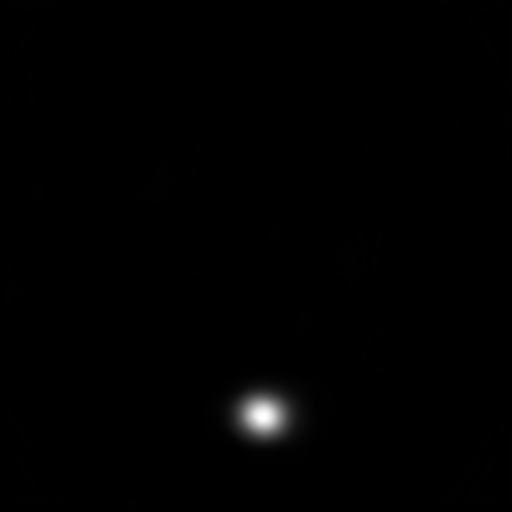}
\includegraphics[width=0.8in,height=0.7in]{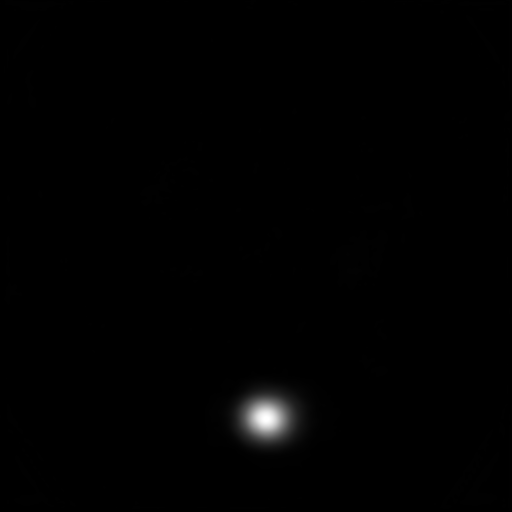}
\includegraphics[width=0.8in,height=0.7in]{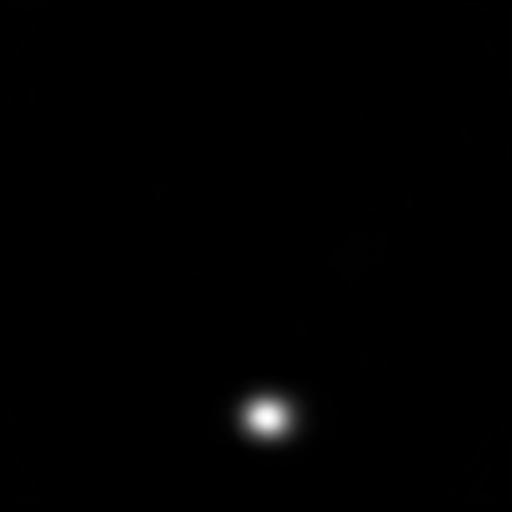}
\includegraphics[width=0.8in,height=0.7in]{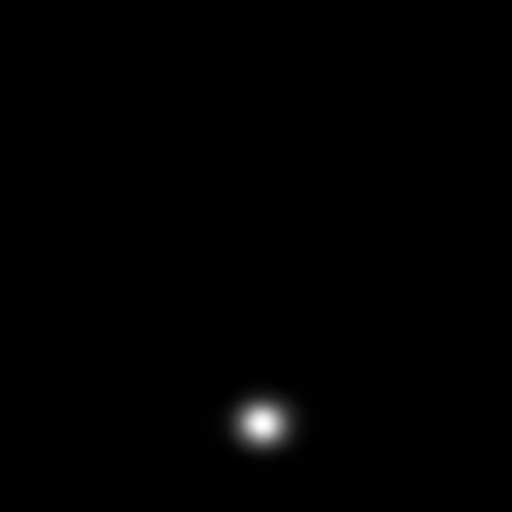}
\includegraphics[width=0.8in,height=0.7in]{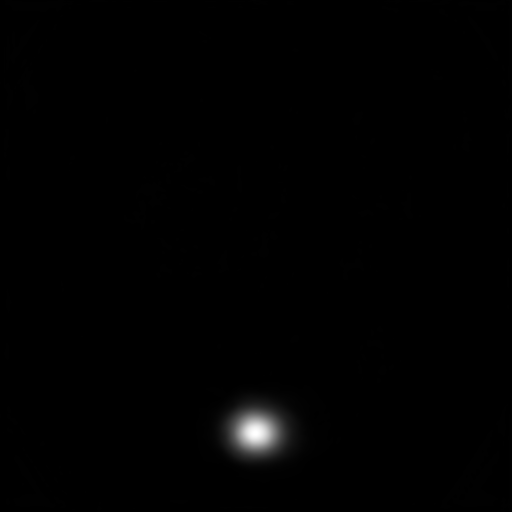}
\includegraphics[width=0.8in,height=0.7in]{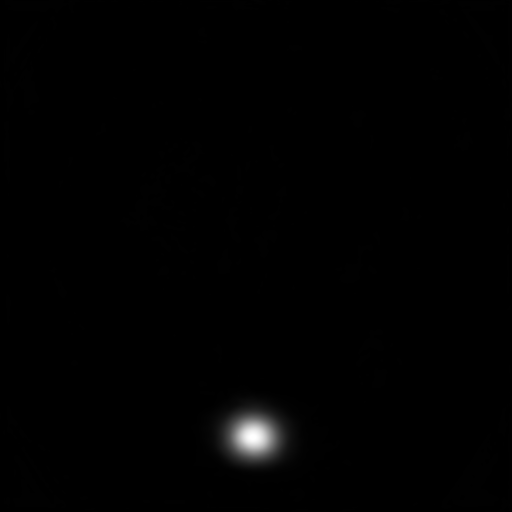}\tabularnewline

\includegraphics[width=0.8in,height=0.7in]{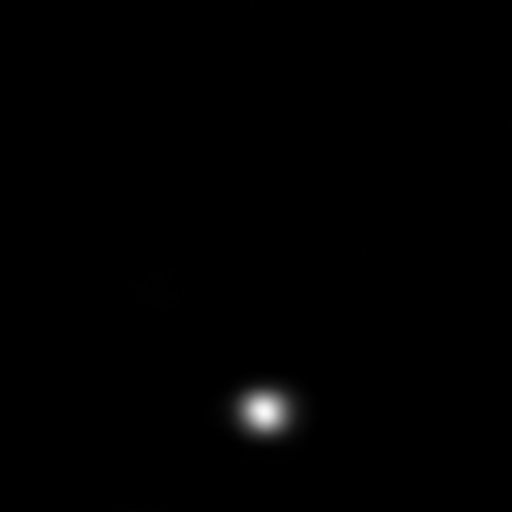}
\includegraphics[width=0.8in,height=0.7in]{figures/series/clean/4/4_frame/221.png}
\includegraphics[width=0.8in,height=0.7in]{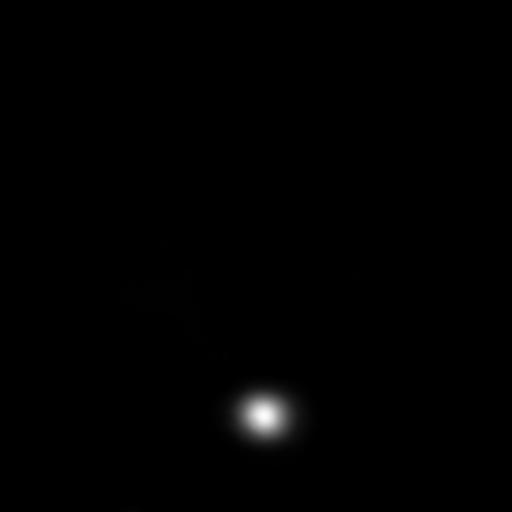}
\includegraphics[width=0.8in,height=0.7in]{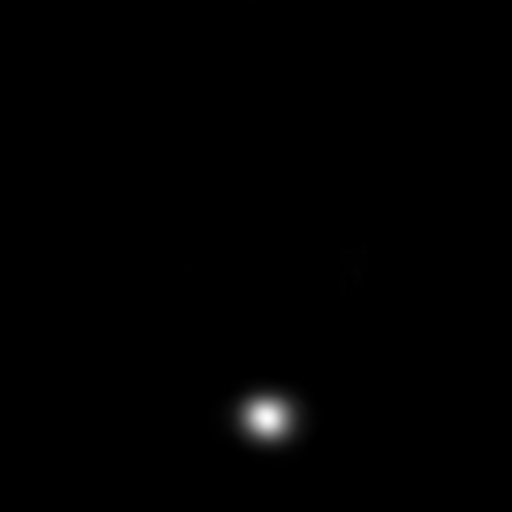}
\includegraphics[width=0.8in,height=0.7in]{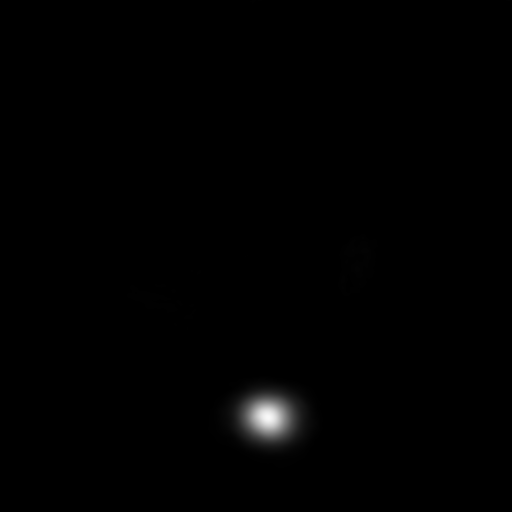}
\includegraphics[width=0.8in,height=0.7in]{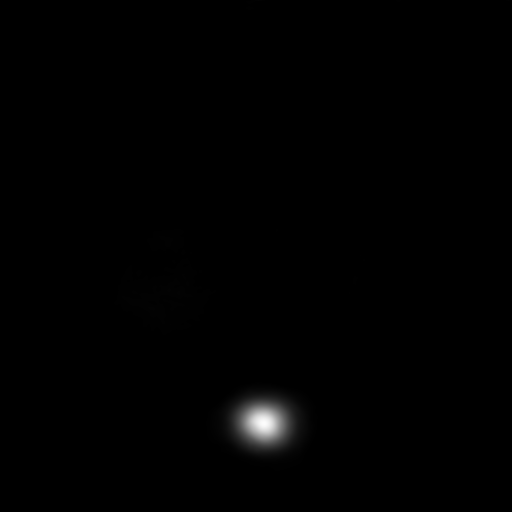}
\includegraphics[width=0.8in,height=0.7in]{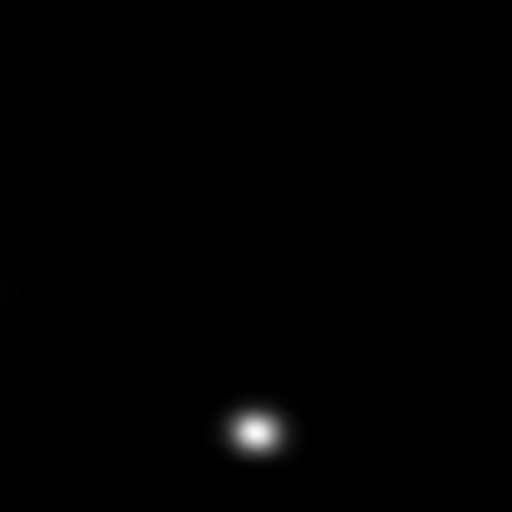}
\includegraphics[width=0.8in,height=0.7in]{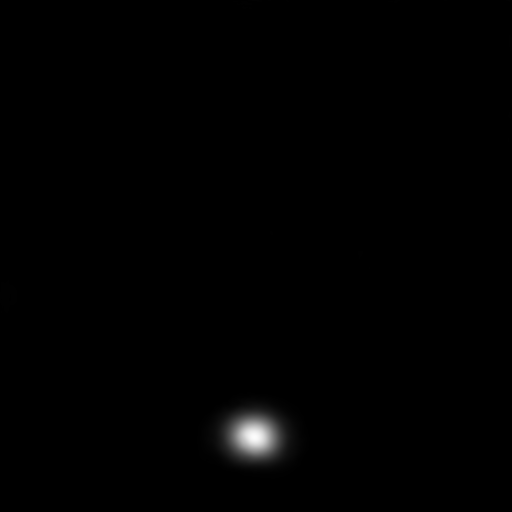}\tabularnewline

\end{tabular}
{\footnotesize\par}}
\par\end{centering}
\caption{An example showing detection outputs of a series of consecutive frames. Each row shows the following:- 1st) input image frames, 2nd) corresponding ground-truth image frames. 3rd) detection output when only the current frame is used, 4th, 5th, and 6th) detection outputs when 1, 2, and 3 previous frames are included in the detection, respectively.}
\label{fig:results}
\end{figure*}
When extracted features from the first previous frame are concatenated with the extracted features of the current frame, the model enjoyed 1.5\% sensitivity increase while the increase in precision is as high as 21.4\%. This result indicates that integrating information from one previous frame can benefit the model to increase both measures by slightly increasing the number of TPs and eliminating a large number of FPs. However, when information from the second and third previous frames is incorporated, model sensitivity degraded while precision improved with a little margin. The resulting degradation in the sensitivity can be due to the relative position of polyps in the current frame w.r.t the farther frames. This occurs when the colonsocopy scope moves fast in the colon leading to dramatic changes in the scene among the consecutive frames, specifically in the farther frames. 


Fig. \ref{fig:results} shows detection outputs in a series of consecutive frames from all four scenarios. As it can be seen, the model obtained much more reliable results from the concatenated features compared to features coming from a single frame. The model generates a large number of FPs when a single frame is under investigation alone.

\subsection{Results on negative videos}
\begin{figure}[htbp]
\centerline{\includegraphics[scale=0.5]{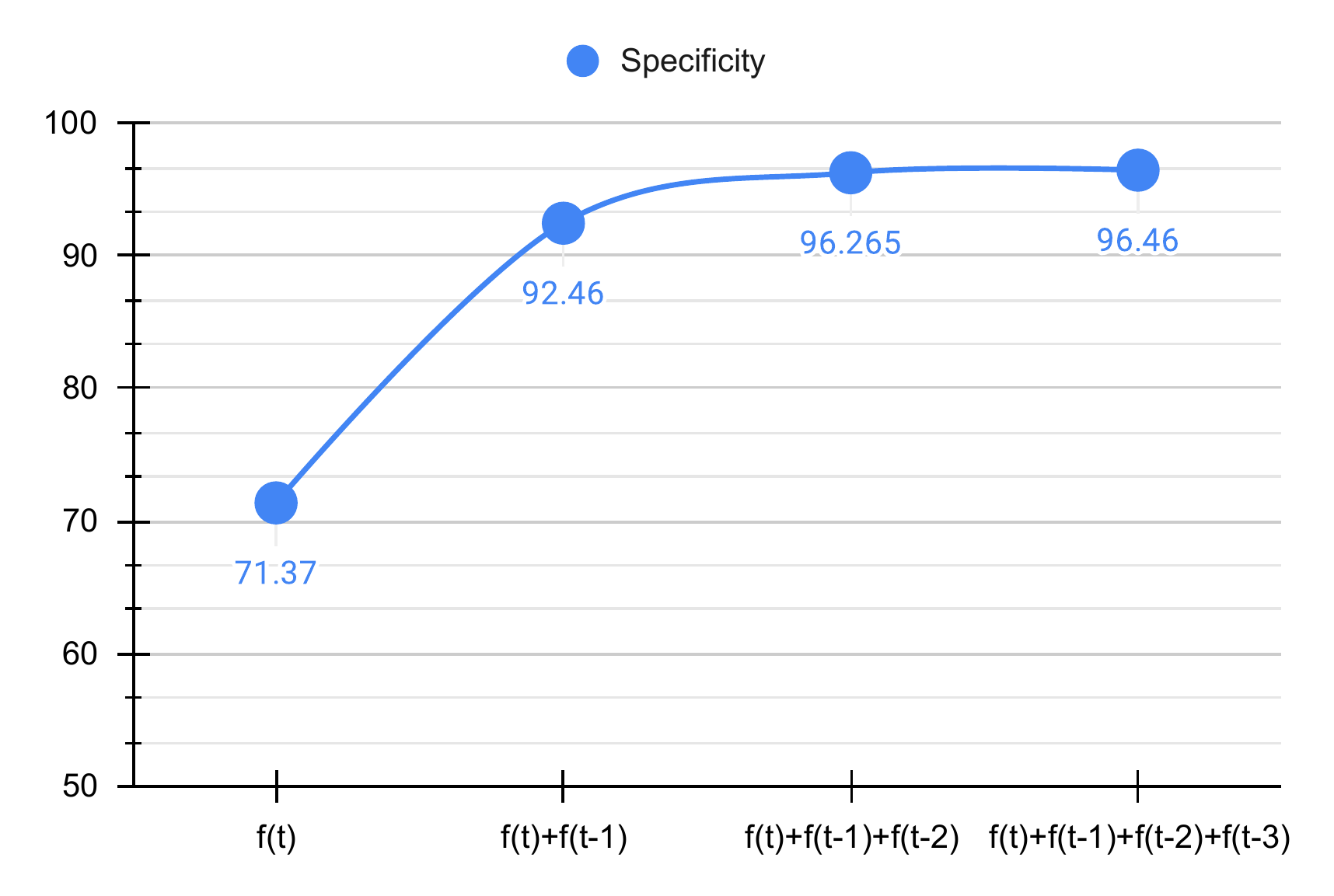}}
\caption{Numerical results showing the performance improvement of the proposed method in terms of specificity}
\label{fig:Spec}
\end{figure}

Fig. \ref{fig:Spec} shows the specificity performance of the model when it was applied to the 10 negative videos. Similar results were observed, i.e., the model's specificity increases when temporal information among neighboring frames is embedded into the model. 

When only the current frame is involved in the detection process, the model specificity is low due to a large number of FPs. When the extracted features of the first previous frame are concatenated with the extracted features of the current frame, the model specificity is raised by 20.09\% which means that a lot of FPs are eliminated. The model continued to increase its specificity when the extracted features of the second and third frames are integrated into the detection process. These results indicate that temporal information is essential for CNNs to potentially reduce the number of FPs and overcome their vulnerability to small changes. 

\begin{figure}[htbp]
\begin{centering}
{\footnotesize{}%
\begin{tabular}{l}
\includegraphics[width=1.05in,height=1.0in]{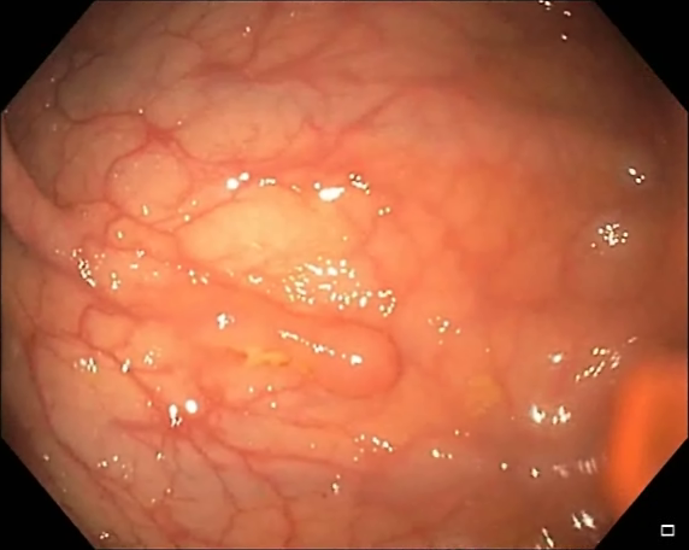}
\includegraphics[width=1.05in,height=1.0in]{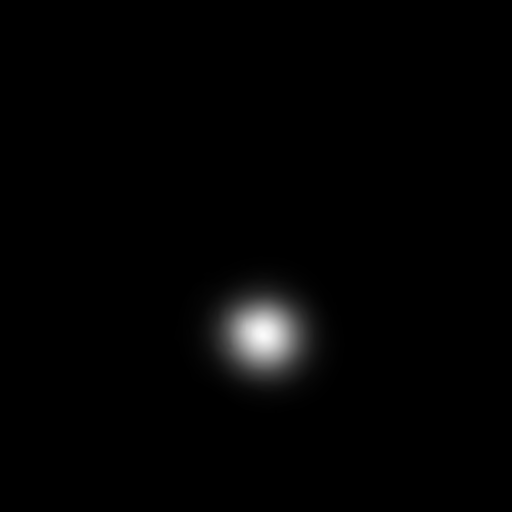}
\includegraphics[width=1.05in,height=1.0in]{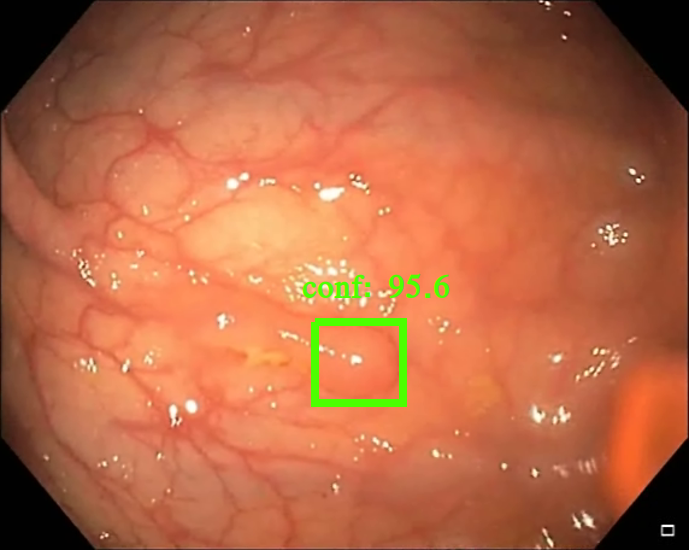}\tabularnewline
\hspace{1.2cm}(a) \hspace{2.3cm} (b) \hspace{2.5cm} (c) \tabularnewline
\end{tabular}
{\footnotesize\par}}
\par\end{centering}
\caption{Final detection output showing the input image (a) and the predicted 2D Gaussian shape (b) projected as a bounding box and confidence value on the input image (c).}
\label{fig:detect}
\end{figure}

We measured the time required by the model to process a single frame. We ran the model on the NVIDIA GeForce RTX 3090 GPU, and we observed the speed of the model which was around 11$\mp$1 msec per frame. It is worth mentioning that the model runs at the same speed in all four scenarios. This is because we equally split the bottleneck layer based on how many previous frames are involved in the process. This way, we avoid increasing the number of activation maps in the bottleneck layer and thus avoid extra mathematical operations. Fig. \ref{fig:detect} presents the final detection output of the proposed method. The predicted 2D Gaussian shapes are projected to bounding boxes and confidence of the detection.  

\section{Conclusions}
In this paper, we presented a novel algorithm to tackle deep learning vulnerability to small changes appearing in neighboring frames. We proposed an efficient method to concatenate extracted CNN features of previous frames with the extracted CNN features of the current frame. We integrated the proposed method into a 2D CNN-based encoder-decoder model because its elegant architecture facilitates this concatenation of feature maps at the bottleneck layer, the latent space, without adding complexity to the model. The obtained results demonstrated that temporal information is essential to improve the overall performance of polyp detection for the analysis of videos. The proposed model was successful to increase the number of true positives and reduce the number of false positives.

\section*{Acknowledgment}
This work was supported partially by the EU project called 5G Health Aquaculture and Transport Validation Trials (5G- HEART) funded by the H2020:ICT framework program under grant agreement number 857034.

\bibliographystyle{IEEEtran}
\bibliography{Reference}

\end{document}